\documentclass[10pt]{article}
\usepackage[a4paper,bindingoffset=0.in,left=2.5cm,right=2.5cm,top=2cm,bottom=3cm]{geometry}
\usepackage{amsmath, amssymb,amsthm}
\usepackage{graphicx,subfigure}
\usepackage{color}
\usepackage{url}
\usepackage[a]{esvect}
\usepackage{enumitem}
\usepackage[utf8]{inputenc}
\usepackage{lmodern}
\usepackage{setspace}
\usepackage{wrapfig}
\usepackage{mdframed}

\usepackage[numbers,sort&compress]{natbib}

\usepackage{amsmath}
\usepackage[capitalize]{cleveref}
\creflabelformat{equation}{#2#1#3}

\newcommand{\edges}{{E}}
\newcommand{\network}{{\mathcal{G}}}
\newcommand{\successors}[1]{{\mathcal{S}(#1)}} 
\newcommand{\Nsuccessors}[1]{{|\successors{#1}|}} 
\newcommand{\data}{{\mathcal{D}}}

\newcommand{\paths}[2]{{ \mathcal{P}_{#1}(#2) }}
\newcommand{\Nnodes}{{|V|}}
\newcommand{\nodes}{{V}}

\newcommand{\AIC}[1]{{\text{AIC}(#1)}}
\newcommand{\BIC}[1]{{\text{BIC}(#1)}}
\newcommand{\EDC}[1]{{\text{EDC}(#1)}}

\newcommand{\MLEmodelMC}[1]{{ \widehat{\mathcal{C}}_{#1} }}
\newcommand{\MLEmodelMON}[2]{{ \widehat{\mathcal{M}}^{#1}_{#2} }}
\newcommand{\eqdef}{\mathrel{\stackrel{\makebox[0pt]{\mbox{\normalfont\tiny def.}}}{=}}}

\newcommand{\MC}[1]{{\mathcal{C}_{#1}}}
\newcommand{\HON}[2]{{\mathcal{N}^{#1}_{#2}}}

\newcommand{\Kgt}{K_\text{gt}}

\newcommand{\dof}[1]{{N_\text{DoF}(#1)}}

\newcommand{\probMC}[2]{{\theta_{#1\rightarrow#2}}}
\newcommand{\probvecMC}[1]{{\vec{\theta}_{#1}}}

\newcommand{\probHON}[2]{{\pi_{#1\rightarrow#2}}}
\newcommand{\probvecHON}[1]{{\vec{\pi}_{#1}}}
\newcommand{\MLEprobHON}[2]{{\hat{\pi}_{#1\rightarrow#2}}}

\newcommand{\probspace}[1]{{ \Delta_{#1} }}

\newcommand{\MON}[2]{{ \mathcal{M}^{#1}_{#2}}}

\newcommand{\obs}[2]{{\mu_\trans{#1}{#2}}}
\newcommand{\obsvec}[1]{{\vec{\mu}_{#1}}}

\newcommand{\hyper}[2]{{\alpha_\trans{#1}{#2}}}
\newcommand{\hypervec}[1]{{\vec{\alpha}_{#1}}}

\newcommand{\trans}[2]{{#1 \rightarrow #2 }}
\newcommand{\Ntrans}[2]{{ n_\trans{#1}{#2} }}
\newcommand{\Ntranstot}{{ n_\text{total}}}

\newcommand{\Betafunc}[1]{B(#1)}
\newcommand{\Dirichlet}[2]{\text{Dir}(#1;#2)}

\newcommand{\euler}{{\mathrm{e}}}

\newcommand{\multiplicative}{{\mathcal{Z}}}
\newcommand{\Kmax}{{K_{\text{max}}}}
\newcommand{\prior}{{\alpha_0}}
\newcommand{\priorK}{{\kappa_0}}

\begin{document}

\title{Learning the Markov order of paths in a network}
\author{
Luka V. Petrovi\'c%
$^{\star}$,
Ingo Scholtes%
$^{\ddagger \star}$,
\\
$^{\star}$  
{\small Data Analytics Group, University of Zurich, Switzerland
}\\
$^{\ddagger}$
{\small
Chair of Data Analytics, University of Wuppertal, Wuppertal, Germany
}
}
\date{}
\maketitle

\newmdenv[hidealllines=true,backgroundcolor=orange!10,skipabove=0pt,skipbelow=0pt,linewidth=.6pt]{codebox}

\begin{abstract}
We study the problem of learning the Markov order in categorical sequences that represent paths in a network, i.e. sequences of variable lengths where transitions between states are constrained to a known graph.
Such data pose challenges for standard Markov order detection methods and demand modelling techniques that explicitly account for the graph constraint.
Adopting a multi-order modelling framework for paths, we develop a Bayesian learning technique that (i) more reliably detects the correct Markov order compared to a competing method based on the likelihood ratio test, (ii) requires considerably less data compared to methods using AIC or BIC, and (iii) is robust against partial knowledge of the underlying constraints.
We further show that a recently published method that uses a likelihood ratio test has a tendency to overfit the true Markov order of paths, which is not the case for our Bayesian technique.
Our method is important for data scientists analyzing patterns in categorical sequence data that are subject to (partially) known constraints, e.g. sequences with forbidden words, mobility trajectories and click stream data, or sequence data in bioinformatics.
Addressing the key challenge of model selection, our work is further relevant for the growing body of research that emphasizes the need for higher-order models in network analysis~\cite{lambiotte2019networks}.
\end{abstract}

\section{Introduction}

Markov chains~\citep{markov1906extension} are a cornerstone in the modelling of categorical sequence data. 
They model a sequence of discrete \emph{states} or \emph{symbols} by means of a discrete-time, stochastic process that has the Markov property, i.e. the next state only depends on the current one.
\citet{markov1913example} first applied Markov chains to the sequence of letters in the poem ``Eugene Onegin'' by Alexander Pushkin.
Today, aside from natural language processing and speech recognition, Markov chain models are used to model sequences in biology, finance, computer science, and network analysis.
Many sequential data in those applications do not satisfy the Markov assumption, i.e. the next state depends not only on the current state but rather on a longer history of states.
To model such data, we can relax the Markov assumption by using Markov chains of higher order, where the order determines the number of previous states that transitions can depend on.

The use of higher-order Markov models raises a long-standing question: What is the optimal order to model a given data set?
Underestimating the optimal order hinders the modelling of relevant patterns in the sequence, reducing the accuracy of predictions and limiting the performance of compression.
Increasing the order $k$ exponentially increases the size of the parameter space of the model, which leads to models that are subject to the curse of dimensionality.
Apart from scalability issues, such models come with a high risk to overfit, to not generalize to new data, and to exhibit poor out-of-sample prediction performance.
Over the last decades, researchers have developed various techniques to address model inference and selection for Markov chains.
However, standard approaches do not incorporate knowledge about \emph{constraints} on transitions, i.e. specific sequences of states that are not possible due to the underlying process.
An important application where such constraints are crucial is the use of Markov chains to model data on \emph{paths in a network}, i.e. sequences of nodes with variable lengths that are traversed in a known graph.
Examples include traces of information propagating in social networks, travel itineraries of passengers in transportation networks, or users navigating hyperlinked pages on the Web.
In such sequences, transitions from one state to other states are constrained by the topology of the network.
Due to recent discoveries of \emph{non-Markovian characteristics} of paths in social networks, transportation, or information systems, the use of higher-order models has become a key approach to develop new network analytic methods for time series data~\cite{lambiotte2019networks}.
The need for techniques to reliably learn the optimal order of Markov chain models of paths in limited amounts of data has been identified as a key challenge in this area.

Despite a large body of inference and model selection techniques for unconstrained categorical sequences~\citep{bartlett1951frequency,hoel1954test,anderson1957statistical,billingsley1961statistical,chatfield1973statistical,akaike1974new,baigorri2009markov,correa2020constrained,csiszar2000consistency,dalevi2005peres,dorea2008optimal,katz1981some,menendez2011testing,moore2018bach,papapetrou2016markov,pethel2014exact,peres2005two,schwarz1978estimating,singer2014detecting,strelioff2007inferring,tong1975determination,van1998testing,zhao2001determination}, ignoring that paths are subject to constraints leads to under-fitting and limits data efficiency~\citep{scholtes2017network}.
The reason why existing methods under-fit can be best understood for techniques relying on AIC and BIC, which account for the degrees of freedom to penalize the complexity of higher-order models.
The topology of a network constrains possible state transitions, which reduces the degrees of freedoms for larger orders.
Standard methods thus over-penalize model complexity and are likely to underestimate the optimal order.

In summary, although the detection of the optimal Markov order for categorical sequential data is a well-studied problem, data on paths pose additional challenges that have not been solved satisfactorily.
Addressing this gap, we use a multi-order modelling framework to derive frequentist and Bayesian model selection techniques for path data.
We experimentally validate them in data with known ground truth and evaluate their data efficiency, i.e. how much data they need to select the correct order.
We repeat our experiment for different maximum model orders, sample sizes as well as for situations where we have only partial knowledge on the network topology.
We find that the Bayesian approach proposed in our work outperforms all other methods for all scenarios, offering substantial improvements in estimation performance and data efficiency.
Regarding a recently published method to address this problem based on the likelihood-ratio test~\citep{scholtes2017network}, we correct the degree of freedom calculation and discover issues that question the Chi-Square approximation of the test statistic.

\section{Background and problem formulation} 
\label{sec:background}

We first introduce order estimation techniques for higher-order Markov models for categorical sequences. 
We then explain additional challenges that occur when modelling paths in networks and specify the learning problem that we address.

\paragraph{Order Estimation for Higher-order Markov chains.}
Consider a categorical sequence $s=(v_1, \ldots, v_l)$ of length $l$, where each symbol or state $v_t$ takes values from alphabet $\nodes$.
A higher-order Markov chain of order $k$ assumes that each symbol $v_t$ is independent of all except the last $k$ symbols, i.e.
\begin{equation}\label{eq:Markov}
  p(v_t| v_1, \ldots, v_{t-1}) = p(v_t|v_{t-k}, \ldots, v_{t-1}) 
\end{equation}
where for $k=1$ we recover an ordinary \emph{memoryless} Markov chain~\cite{markov1906extension}.
We use the term ``zeroth-order Markov chain'' to refer to a model where symbols $v_t$ are i.i.d.
For a given $k$-th order Markov chain with a specific choice of parameters that we denote as $\MC{k}$, the conditional probability of symbol $v_t$ for a given \emph{history} of symbols $(v_{t-k}, \ldots, v_{t-1})$ is
\begin{equation}\label{eq:MCparams}
  p(v_t|(v_{t-k}, \ldots, v_{t-1}), \MC{k}) = \probMC{(v_{t-k}, \ldots, v_{t-1})}{v_t}
\end{equation}
where $\MC{k}$ consists of all transition probabilities $\probMC{(v_{i-k}, \ldots, v_{i-1})}{v_i}$.
Organizing $\probMC{h}{v}$ by histories $h$, we can bring them in vector-form as:
\begin{align}
  \probvecMC{h} \eqdef ( \probMC{h}{v_1}, \ldots, \probMC{h}{v_\Nnodes} ) \text{ such that } \forall h: \sum_{v} \probMC{h}{v} = 1 \wedge \forall v: 0 \leq \probMC{h}{v} \leq 1
  \label{eq:MCparamSpace}
\end{align}
For each history $h$, $\probvecMC{h}$ is a point in a \emph{probability simplex} $\probspace{\Nnodes}$ in $\Nnodes$-dimensional space, which has $\Nnodes-1$ degrees of freedom due to the normalization.
The parameter space of a $k$-th order Markov chain is then the Cartesian product of $\probspace{\Nnodes}$ over all histories $h$ of length $k$, ($h\in\nodes^k$):
\begin{equation}
  \MC{k} \in \prod_{h\in\nodes^k} \probspace{\Nnodes}
\end{equation}
Therefore, a Markov chain of order $k$ has $\dof{\MC{k}} = \Nnodes^k(\Nnodes-1)$ degrees of freedom.

For a given sequence $s$ and fixed order $k$ we can, e.g., use likelihood maximization to learn the parameters $\MLEmodelMC{k}$ of the $k$-th order Markov chain that best ``explains'' patterns in the sequence $s$.
Higher-order Markov chains with different orders are nested, i.e. for any choice of parameters $\MC{k}$ of a $k$-th order model, a point $\MC{k'}$ in the parameter space of a model with order $k'>k$ exists such that the likelihoods of the two models are identical.
This implies that a maximization of model likelihood trivially chooses the largest order available, since a model with smaller order can never have larger likelihood.
Markov order detection is thus an exemplary instance of a model selection problem in which need to account both for the goodness-of-fit (e.g. expressed in the likelihood) and model parsimony (e.g. expressed in the degrees of freedom).
A number of methods have been developed for this long-standing problem, including methods based on various estimators \cite{baigorri2009markov,peres2005two,dalevi2005peres,zhao2001determination,dorea2008optimal}, surrogate data \cite{van1998testing,pethel2014exact,correa2020constrained}, generalizations of a likelihood ratio test \cite{menendez2011testing}, or mutual information \cite{papapetrou2013markov,papapetrou2016markov}.

Some of the earliest works have applied a likelihood-ratio test \cite{bartlett1951frequency,anderson1957statistical,billingsley1961statistical,chatfield1973statistical} to determine the optimal Markov order of a sequence.
Considering that higher-order Markov chains with orders $k$ and $k'>k$ are nested models, and assuming that the sequence $s$ is sufficiently long, Wilks' theorem allows us to approximate the distribution of the test statistic based on the $\chi^2$-distribution, where the parameter of the distribution is the difference $\dof{\MC{k'}}-\dof{\MC{k}}$ between the degrees of freedom of a $k'$- and a $k$-th order model.
We can use this to calculate a $p$-value of the null hypothesis that the fitted parameter $\MLEmodelMC{k'}$ is in the subset of the model parameters which correspond to a $k$-th order Markov chain.
This approach naturally accounts for the expected increase in model likelihood that is due to the larger degrees of freedom of a Markov chain with order $k'>k$.
\citet{tong1975determination} applied \emph{Akaike's information criterion} (AIC)~\citep{akaike1974new} to Markov order detection. 
This criterion balances the degrees of freedom $\dof{\MC{k}}$ of a $k$-th order Markov chain with the goodness-of-fit captured in the likelihood $p(s|\MLEmodelMC{k})$.
\citet{schwarz1978estimating} proposed the \emph{Bayesian information criterion} (BIC), which is based on the asymptotic behavior of Bayes estimators. 
Similar to the AIC, it incorporates both the goodness-of-fit and the degrees of freedom of the model.
\citet{katz1981some} showed that AIC is an inconsistent estimator, prone to over-fitting even at large sample sizes, while BIC has been proven to be consistent \cite{csiszar2000consistency}.
Most relevant for our work is the Bayes factor method \citep{kass1995bayes} that is based on Bayes rule.
It was used by \citet{liu1999bayesian,strelioff2007inferring,singer2014detecting} to determine the optimal order of Markov chain models for general categorical sequences.

\paragraph{Problem statement.}
Different from the works above, we address the problem of learning the optimal Markov order based on data of paths in a known network.
Examples for such data include click streams generated by multiple sessions of users that navigate a hyperlinked document graph, collections of ticket data that capture the flight itineraries of passengers in an airline network, or information propagating along the edges of a social network.
We assume that the data $\data$ is an unordered \emph{multiset of constrained, variable-length sequences} that correspond to \emph{paths in a given network} $\network = (\nodes, \edges)$ with nodes $\nodes$ and edges $\edges \subseteq \nodes \times \nodes$.
A path in a network $\network$ is a sequence of nodes $(v_1, v_2, \ldots, v_l)$, $v_t \in \nodes$, where consecutive nodes are connected by an edge, i.e. $ \forall t: (v_t, v_{t+1}) \in E$.\footnote{We do not require nodes or edges to be unique, i.e. we do not distinguish between paths, trail, or walks~\cite{bollobas2013modern}.}
We denote the set of successors of node $v$ as $\successors{v}=\{ w \in \nodes: (v, w) \in \edges \}$ and successors of a path $h = (h_1, \ldots, h_l)$ as $\successors{h} = \successors{h_l}$.
We denote the set of all paths of length $k$ in a network $\network$ as $\paths{\network}{k}$.
To model such data, researchers in network science have studied \emph{higher-order network models}~\cite{lambiotte2019networks}, which combine higher-order Markov chains with network-based models for paths.
Like higher-order Markov chains, a higher-order network (HON) with order $k$ assumes a $k$-th order Markov property (\cref{eq:Markov}).
However, for inference and model selection tasks, it explicitly accounts for those state transitions that correspond to a possible path in a given network $\network$.
For a given $k$-order HON on $\network$ with parameters that we denote as $\HON{\network}{k}$, the transition probabilities are
\begin{align}
  p(v_i|(v_{i-k}, \ldots, v_{i-1}), \HON{\network}{k}) = \probHON{(v_{i-k}, \ldots, v_{i-1})}{v_i}, \text{where } \forall  j \in \{i-k, \ldots, i\}: v_j \in \successors{v_{j-1}}.
  \label{eq:HONparams}
\end{align}
Again grouping parameters by histories, we obtain:
\begin{align}
  &\probvecHON{h} \eqdef ( \probHON{h}{v_1}, \ldots, \probHON{h}{v_{|\successors{h}|}} )  \in \probspace{|\successors{h}|}\\
  &\HON{\network}{k} \in \prod_{h\in \paths{\network}{k}} \probspace{ \Nsuccessors{h}}
  \label{eq:HONparamSpace}
\end{align}

The difference in modelling a large number of (typically) short paths and a single long sequence might look small, but it poses a major issue.
Neither Markov chains nor HONs of order $k$ capture probabilities for histories shorter than $k$.
For a single long sequence omitting the first $k$ symbols might be negligible.
However, for data with a large number of short paths we would omit the first $k$ symbols of \emph{each} path.
Instead of a single HON, \citet{scholtes2017network} proposes to combine multiple HONs of orders $k \in \{0, 1, \ldots, K\}$ as ``layers'' of a ``multi-order network'' (MON) model $\MON{\network}{K} = \{\HON{\network}{k}\}_{k = 0}^{K}$.
Similar to a Markov chain of order $K$, a MON with maximal order $K$ assumes a $K$-th order Markov property (\cref{eq:Markov}), modelling all transition probabilities for histories longer than $K$ with the $K$-th order layer.
Transition probabilities for histories of length $k<K$ are modelled with the $k$-th layer.
The probability of a sequence $s = (v_0, \ldots, v_l)$ for a MON model $\MON{\network}{K} = \{\HON{\network}{k}\}_{k = 0}^{K}$ is:
\begin{equation}
  p(s|\MON{\network}{K}) = \prod_{i<K} p(v_i| (v_0, \ldots, v_{i-1}), \HON{\network}{i}) \prod_{K\leq i \leq l} p( v_i|(v_{i-K},\ldots, v_{i-1}), \HON{\network}{K}) 
  \label{eq:MONlikelihood}
\end{equation}
Since it combines multiple HONs, a MON model has $\dof{\MON{\network}{K}} = \sum_{k = 0}^K\sum_{h\in \paths{\network}{k}} \Nsuccessors{h}-1$ degrees of freedom. 
With $\paths{\network}{0} \eqdef \{\epsilon\}$ for the empty history $\epsilon$, its parameter space is:
\begin{equation}
  \MON{\network}{K} \in \prod_{k=0}^{K} \prod_{h\in \paths{\network}{k}} \probspace{ \Nsuccessors{h}}
  \label{eq:MONparamSpace}
\end{equation} 
In the remainder of this work we address the problem of learning the optimal maximum order $K$ of a multi-order network model, given paths $\data$ and a network $\network$.

\section{Learning the Markov order of paths} 
\label{sec:methods}

We introduce four model selection techniques to learn the optimal maximum Markov order for a multi-order network model of paths.
We first derive a Bayesian method to learn both the optimal parameters of a multi-order network model as well as the Markov order of paths using Bayes factors~\citep{kass1995bayes}.
We then adapt three MLE-based Markov order estimation techniques to our problem.

\paragraph{Bayesian learning of model parameters and optimal maximum Markov order.}
We say that, in a path $s = (v_1, \ldots, v_l)$ every node $v_t$ represents one observation of a transition from history $h$ to node $v_t$, where $h = (v_1, \ldots, v_{l-1})$.
A single path $s$ of length $|s|=l$ contains transitions for histories with multiple lengths, from length zero $\trans{\epsilon}{ v_1}$ (with empty history $\epsilon$) to $l-1$ $\trans{( v_1, \ldots, v_{l-1})}{v_l}$. 
We denote the number of observed transitions from history $h$ to node $v$ in a data set of paths $\data$ as
\begin{equation}
  \Ntrans{h}{v} \eqdef 
  \sum_{s\in\data} 
    \sum_{ 1\leq t\leq|s|} 
      \delta(v,  v_t)
      \delta((v_1, \ldots, v_{t-1}), h)
  ,
  \text{ with } 
  \delta(x,y) \eqdef 
  \begin{cases}
    1, x=y,\\ 
    0, x\neq y
  \end{cases}
\end{equation}

For each given maximum order $K$, the Bayesian approach keeps track of the probability density $p(\MON{\network}{K}|K, \network)$ over the parameter space of a multi-order model.
Model parameters $\MON{\network}{K}$ consist of transition probabilities $\probvecHON{h}$ to nodes from $\successors{h}$ for all histories $h$.
We can model the probability density of $\probvecHON{h}$ by a Dirichlet distribution~\citep{liu1999bayesian,strelioff2007inferring} with concentration (hyper)parameters $\hypervec{h}$ ($\Betafunc{\vec{x}}$ denotes the well-known multivariate Beta function):
\begin{align}
  p(\probvecHON{h}) &= \Dirichlet{\probvecHON{h}}{\hypervec{h}} = \frac{1}{\Betafunc{\hypervec{h}}} \prod_{v_i \in \successors{h}} \probHON{h}{v_i}^{\hyper{h}{v_i}-1}\\
  \hypervec{h} &\eqdef (\hyper{h}{v_1}, \hyper{h}{v_2},\ldots,\hyper{h}{v_{|\successors{h}|}}), v_i\in \successors{h}
\end{align}
We denote the prior with $\prior$ and use $\prior = 1$ for a so-called "flat" prior, where the probability density is constant over the parameter space. 
This corresponds to choosing all hyperparameters as one, i.e.
\begin{equation}
  p(\MON{\network}{K}|\prior=1, K, \network) = \prod_{k=0}^{K} \prod_{h\in \paths{\network}{k}} \Dirichlet{\probvecHON{h}}{\vec{1}_\Nsuccessors{h}} = \text{\it const.}
  \label{eq:prior}
\end{equation}

From Bayes rule, we obtain the following update rule for the distribution of model parameters $\MON{\network}{K}$ with maximum order $K$ on a network $\network$ given path data $\data$ and prior $\prior$:
\begin{equation}
  \label{eq:update}
  p(\MON{\network}{K}|\data, \prior, K, \network) = \frac{p(\data|\MON{\network}{K}, \prior, K, \network) p(\MON{\network}{K}|\prior, K, \network)}{p(\data|\prior, K, \network)}
\end{equation}
For given parameters $\MON{\network}{K}$ of a MON model of order $K$ on $\network$, the likelihood does not depend on the choice of the prior, i.e. $p( \data| \MON{\network}{K}, \prior, K, \network) = p(\data|\MON{\network}{K}, K, \network)$. 
Using the $K$-th order Markov property (\cref{eq:Markov}), we can compute the likelihood of parameters $\MON{\network}{K}$ given data $\data$ as
\begin{equation}
  p(\data|\MON{\network}{K}, K, \network)
    = 
      \multiplicative
      \prod_{k = 0}^{l_{\text{max}}}
      \prod_{ h\in  \paths{\network}{k}} 
      \prod_{v \in \successors{h}}
        p(\Ntrans{h}{v}|\MON{\network}{K}, K, \network) 
    = 
      \multiplicative
      \prod_{k = 0}^{K}
      \prod_{ h\in \paths{\network}{k}} 
      \prod_{v \in \successors{h}}
        \probHON{h}{v}^\obs{h}{v}
    \label{eq:likelihood} 
\end{equation}
where $\multiplicative$ is the number of permutations of paths in $D$, $l_{\text{max}}$ is the maximal length of paths in $\data$, and $\obs{h}{v}$ counts the transitions in the data assuming the $K$-th order Markov property as follows:
\begin{equation}
  \obs{h}{v} = 
  \left\{
  \begin{aligned}
    &\Ntrans{h}{v}, 
    &\text{for } |h|<K\\
    \sum\limits_{\substack{x \in \paths{\network}{L},  L\geq K}} 
      &\Ntrans{x}{v}
      \delta((x_{L-K+1}, \ldots, x_L), h), 
    &\text{for } |h|= K
  \end{aligned}
  \right.
  \label{eq:evidence}
\end{equation}
We organize $\obs{h}{v}$ as vectors $\obsvec{h} = (\obs{h}{v_1}, \ldots, \obs{h}{v_{\Nsuccessors{h}}})$.
We then use \cref{eq:likelihood}, and the moments of the Dirichlet distribution to derive the marginal likelihood (derivation in \cref{sec:Marginal likelihood: step by step derivation}):
\begin{align}
  p(\data|\prior, K, \network) 
    &= 
      \int\limits_{\prod_{h} \probspace{ \Nsuccessors{h}} } 
        p(\data|\MON{\network}{K}, \prior, K) p(\MON{\network}{K}|\prior, K) d\MON{\network}{K}
    = 
    \multiplicative
      \prod_{k = 0}^{K}
      \prod_{ h\in \paths{\network}{k}}
      \frac{
        \Betafunc{\hypervec{h}+\obsvec{h}}
      }{
        \Betafunc{\hypervec{h}}
      }  
    \label{eq:normalization}
\end{align}
Substituting back into \cref{eq:update}, we arrive at the simple update rule:
\begin{equation}
  \hypervec{h}^\text{posterior} = \hypervec{h}^\text{prior}  + \obsvec{h}
\end{equation}

This yields a Bayesian method to infer parameters of a multi-order model based on a large collection of paths $\data$ and a fixed maximum order $K$. 
To learn the optimal order $K$ we again apply Bayes rule.
We first choose the orders that we want to compare, e.g., $K \in\{0, 1,\ldots, \Kmax\}$ up to a maximum value of  $\Kmax$.
We denote the choice of the prior over $K$ as $p(K| \priorK) = \priorK(K)$.
For the inference of the optimal $k$ order for $k$-th order Markov chain models of unconstrained sequences (rather than multi-order models of paths), \citet{strelioff2007inferring} mention two common priors: a uniform prior and a prior that additionally penalizes more complex models.
Due to its superior performance for our problem, we limit our discussion to the uniform prior (see results for the latter prior in \cref{sec:Exponential penalization of higher orders}):
\begin{equation}
  \priorK(K) \eqdef (\Kmax+1)^{-1} = \text{\it const.}
  \label{eq:uniformPrior}
\end{equation}
For a data set $\data$ and order $K$, we use Bayes rule to calculate posterior probabilities:
\begin{equation}
  p(K|\prior, \priorK, \data, \network) = 
    \frac{
      p(\data|\prior, \priorK, K, \network) 
      p(K|\prior, \priorK, \network)
    }{
      \sum_{K\in \mathbb{K}}
        p(\data|\prior, \priorK, K, \network)
        p(K|\prior, \priorK, \network)
    }
    = \frac{
      p(\data|\prior, K, \network) 
      \priorK(K)
    }{
      \sum_{K\in \mathbb{K}}
        p(\data|\prior, K, \network)
        \priorK(K)
    }
  \label{eq:bayesK}
\end{equation}
which we calculate analytically from \cref{eq:normalization} since the model evidence is equal to marginal likelihood.
The ratio of probabilities $B_{K,K'} = \frac{p(K'|\prior, \priorK, \data, \network)}{p(K|\prior, \priorK, \data, \network)}$ is the Bayes factor.
To facilitate the comparison between Bayesian and MLE-based model selection, we use hypothesis tests from \cite{kass1995bayes} to output a single order, instead of assigning probabilities to each order.
We chose the maximal $K'$ that is significantly more likely than all models with $K < K'$.
We use significance levels from \cite{kass1995bayes}, i.e. we find ``positive'' evidence in favor of $K'$ over $K$ iff $B_{K,K'}>3$ and ``very strong'' evidence iff $B_{K,K'}>150$.
Since likelihoods are marginalized over the parameter space, larger $K$ do not necessarily have larger likelihood.
This naturally introduces Occam's razor~\citep{mackay2003information} and avoids overfitting.

\paragraph{MLE-based Markov order estimation}
Apart from the Bayesian method above, we can use maximum likelihood estimation (MLE) to infer transition probabilities based on the paths observed in a data set $\data$.
We obtain an MLE of parameters $\MLEmodelMON{\network}{K}$ as a ratio of transition frequencies 
\begin{equation}\label{eq:MLE}
  \MLEprobHON{h}{v} = \frac{ \obs{h}{v} }{ \sum_{w \in \nodes} \obs{h}{w}}.
\end{equation}
To learn the optimal order $K$ we can use information criteria like AIC or BIC, which for a multi-order model with estimated parameters $\MLEmodelMON{\network}{K}$ are given as:
\begin{align}
  \label{eq:AIC}
  \AIC{K} & \eqdef -2\log{p(\data| \MLEmodelMON{\network}{K})} + 2\dof{\MLEmodelMON{\network}{K}}\text{,} \\
  \BIC{K} & \eqdef -2\log{p(\data| \MLEmodelMON{\network}{K})} + \dof{\MLEmodelMON{\network}{K}}\log{\Ntranstot}\text{.}
\end{align}
where $\Ntranstot \eqdef \sum_{h, v} \Ntrans{h}{v}$ is the total number of transitions on all paths in $\data$ and $\dof{\MON{\network}{K}}$ are the degrees of freedom of a multi-order model. 
Note that $\dof{\MON{\network}{K}}$ depends on the maximum order $K$ and the topology of the network $\network$, where sparser networks typically lead to smaller degrees of freedom.
While we omit it due to space constraints and inferior performance, in \cref{sec:Efficient determination criterion (EDC)} we additionally adapt the EDC~\cite{zhao2001determination} to multi-order models.

We finally discuss a method to learn the optimal maximum order of a multi-order network model based on a likelihood-ratio test, which was previously adapted in~\cite{scholtes2017network}.
For two candidate orders $K'>K$, we define the test statistic based on the ratio of model likelihoods as:
\begin{equation}
  x 
  \eqdef
    -2\log{
    (
      p(\data| \MLEmodelMON{\network}{K} ) / 
       p(\data| \MLEmodelMON{\network}{K'})
    )
      }
  \end{equation}
Due to the nestedness of models (\cref{sec:background}), $x$ approximately follows a Chi-square distribution, i.e.
\begin{equation}
  p(x) \approx \chi^2(x; \xi); \text{ with } \xi = \dof{\MON{\network}{K'}}-\dof{\MON{\network}{K}}\text{.}
  \label{eq:chi-squared approximation}
\end{equation}
where $\dof{\MON{\network}{K}}$ are the degrees of freedom of a multi-order model with maximum order $K$.
As explained in \cref{sec:background}, we can learn the optimal order by calculating a $p$-value of the null hypothesis that the observed increase in likelihood is due to the additional complexity of a model with larger oder $K$.
While the same idea was used in \cite{scholtes2017network}, we highlight two important differences:
First, for the degrees of freedom \cite{scholtes2017network} does not distinguish between the underlying network and observed transitions in paths, i.e. it relies on the assumption that every possible edge in the network is traversed at least once.
Second, despite a correct explanation in the text, there is a mistake in Eq. (8) and (9) of \citep{scholtes2017network} that leads to an overestimation of the degrees of freedom, which we corrected in \cref{sec:background}.

\section{Experimental analysis} 
\label{sec:experiments}

\begin{figure}[!ht]
  \centering
  \includegraphics[width = .9\linewidth]{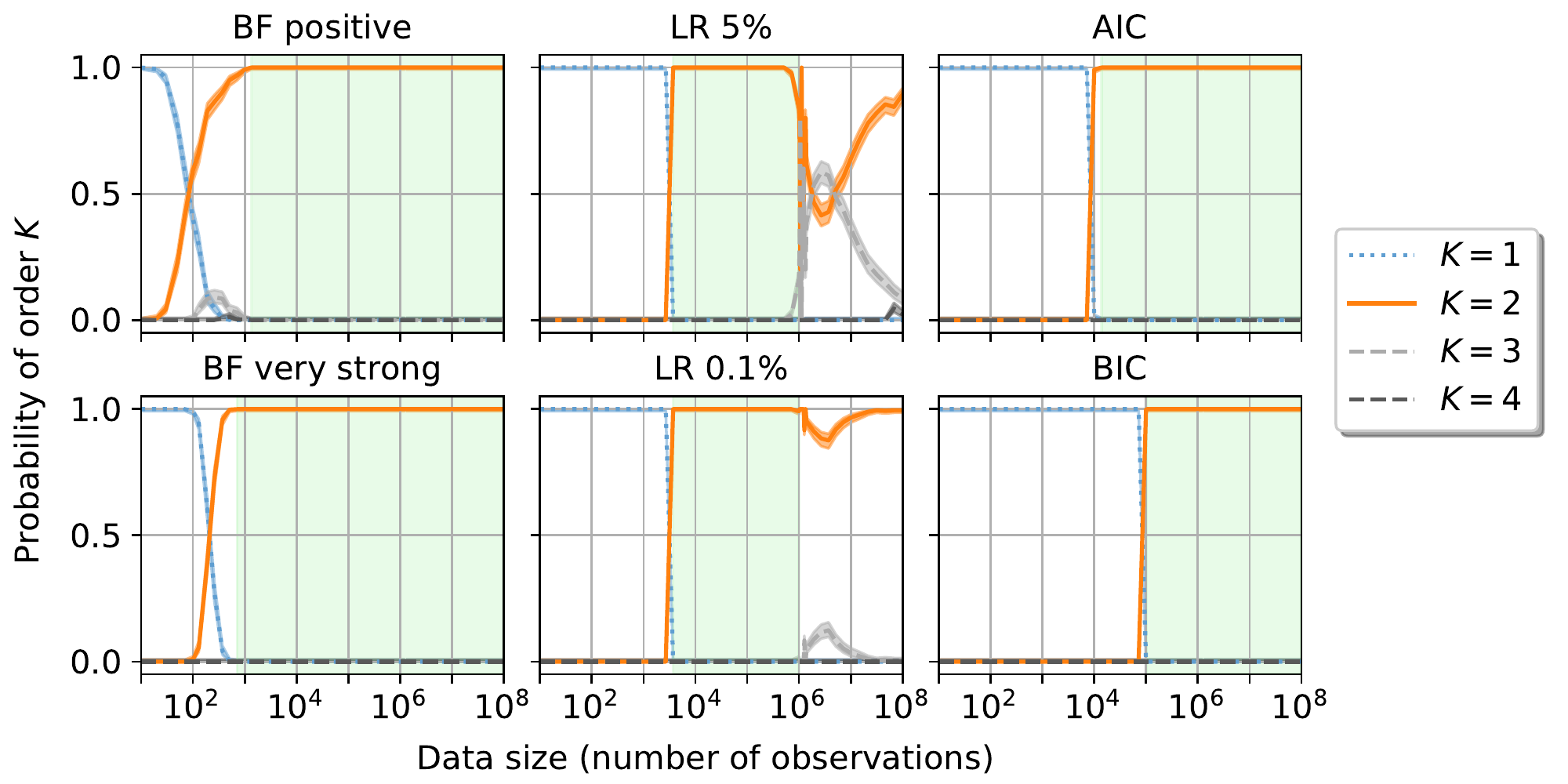}
  \caption{
  We experimentally evaluate order estimation using AIC, BIC, likelihood ratio test (LR) with significance thresholds $5\%$ and $0.1\%$, and Bayes factors (BF) with ``positive'' and ``very strong'' evidence~\cite{kass1995bayes} for paths with a ground truth maximum order $\Kgt = 2$.
  We show the frequency of choosing order $K$ (y-axis) for different data sizes (x-axis).
  For each data size we ran $500$ independent experiments, each generating a random graph with $n=100, m = 350$ (and removing nodes with no successors) as well as a multiset of random paths in this network.
  Confidence intervals around curves are Wilson score intervals with $95\%$ coverage. Shaded areas indicate data sizes where methods learned the correct order in all $500$ experiments.
  }
  \label{fig:order detection comparison}
\end{figure}

We now experimentally evaluate the model selection techniques introduced above in synthetic paths with known ground truth order.
For a ground truth maximum order $\Kgt$, we first generate a random undirected and unweighted graph $\network$.
We then choose a multi-order network (MON) model $\MON{\network}{\Kgt}$ uniformly at random from the space of possible models (\cref{eq:MONparamSpace}).
We finally use that model to generate a set of paths $\data$ with a given size $\Ntranstot$.
Assuming that we have access to the path data $\data$ and the network $\network$ constraining paths, we construct MON models with different maximum orders $K$, fit them to the paths $\data$, and determine the optimal maximum order using each of the four model selection techniques discussed in \cref{sec:methods}.
We repeat the procedure above for $500$ random networks and determine the rates at which the methods select each order $K$ as optimal.
We repeat this experiment for different data sizes $\Ntranstot$.
In \cref{fig:order detection comparison} we present results for a random graph with $100$ nodes in $350$ edges and paths with ground truth order $\Kgt = 2$ (analogous figure for $\Kgt=3$ in \cref{sec:other_results}).
The figure shows the frequencies ($y$-axis) at which the methods select order $K\ \in \{ 1, \ldots, 4\}$ (four curves) for data with varying size $\Ntranstot$ ($x$-axis).
Shaded areas in \cref{fig:order detection comparison} indicate the range of data sizes where the correct order was detected in all $500$ experiments (AIC, BIC, BF) or, for  the likelihood ratio test, where the observed type-I error rate was smaller than the significance threshold.

\begin{figure}[!ht]
  \centering
  \includegraphics[width = .9\linewidth]{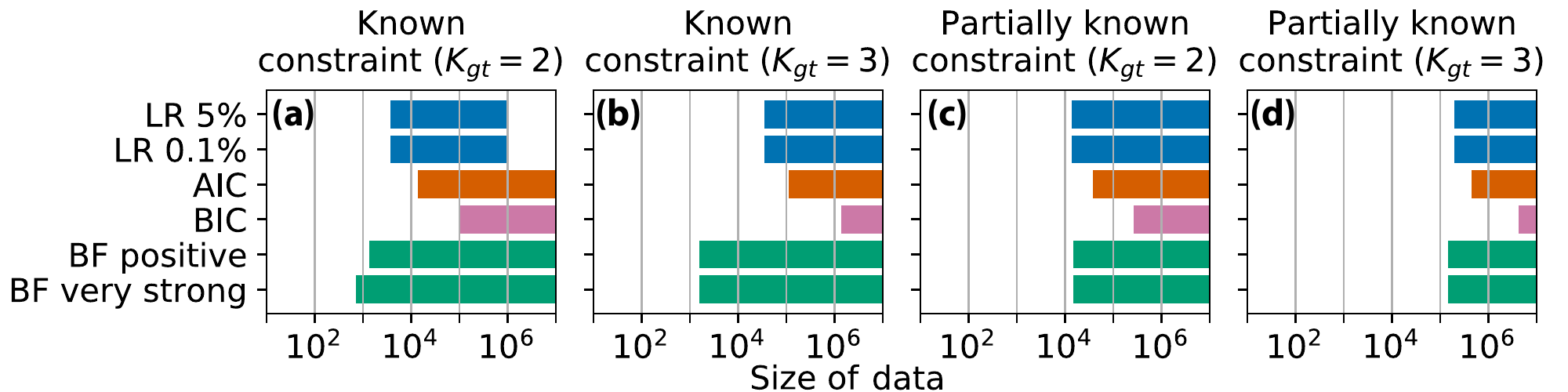}
  \caption{
  Ranges of data size where the methods learned the correct order in all $500$ repetitions. Experimental setup is identical to \cref{fig:order detection comparison}, i.e. bars in (a) correspond to shaded areas in \cref{fig:order detection comparison}. In (b) we set the ground truth $K_{gt}=3$, in (c-d) we reduce knowledge about the network constraint.
  }
  \label{fig:correct order detection ranges}
\end{figure}

The results in \cref{fig:order detection comparison} indicate a strong difference between the methods in terms of data efficiency, i.e the minimal number of observations needed to reliably detect the ground truth order.
To further highlight this aspect of data efficiency, in \cref{fig:correct order detection ranges} we compare the ranges of data sizes for which the four methods determine the correct order.
\cref{fig:correct order detection ranges} (a) shows these ranges for $\Kgt = 2$ (i.e. the same experiments presented in \cref{fig:order detection comparison}), while \cref{fig:correct order detection ranges} (b) shows the results for $\Kgt = 3$.

\newcommand{\impnetwork}{{\mathcal{G}'}}
\newcommand{\fakeEdges}{{E'}}

All experiments above assume that we have full information on the network that constrains which paths are possible.
However, in real data we often have partial information about the underlying network.
For instance, we might have access to the network of streets in a city, but lack information on road signs that indicate forbidden turns and one-way lanes.
To simulate this situation, we add random edges to the network $\network$ used to generate paths, i.e. we use $\impnetwork = (\nodes, \edges \cup \fakeEdges)$ to constrain the model, where $\fakeEdges$ denote transitions that are apparently allowed, but cannot be realized in the actual network $\network$.
We show the ranges of data size where each method detects the correct order for the partially known constraint in panels (c) and (d) in \cref{fig:correct order detection ranges}.
These experiments are performed with the same parameters as above, adding $350$ additional directed edges to the network.

\paragraph{Discussion}
In \cref{fig:order detection comparison} we note that the likelihood ratio test overfits the Markov order of paths.
More precisely, using the $\chi^2$-approximation of the test statistic, the type-I error rate converges to the chosen significance threshold $p_{\text{thres}}$ \emph{from above}.
The type-I error rate is zero for small data sizes, exceeds the significance threshold $p_{\text{thres}}$ around $10^6$ observations, and as the data-size increases, slowly approaches $p_{\text{thres}}$.
For moderate data sizes, our results in \cref{fig:order detection comparison} further indicate that the likelihood ratio test is prone to make type-I errors with a probability that is orders of magnitude larger than $p_{\text{thres}}$. 
The curve of $K = 3$ shows the frequency with which order three is chosen rather than the correct order $\Kgt=2$, which is a \emph{lower bound} for the type-I error rate.
For the chosen significance threshold $p_{\text{thres}}=0.05$, we find that the probability to make a type I error exceeds $0.5$, while for $p_{\text{thres}}=0.001$ it exceeds $0.1$.
Despite having a small network and a large number of observations, for $p_{\text{thres}}=0.05$ and ground truth $\Kgt=2$ the likelihood ratio test even chose order $4$ with frequency larger than $0.04$, which we expect to happen with probability $0.05\times 0.05 = 0.0025$.
This tendency of the likelihood ratio test to overfit did not show in the experiments with partial knowledge of the constraint. 
We suspect that the overestimation of the degrees of freedom that is due to the partial knowledge masks the tendency to overfit.
Due to this overcounting, the $\chi^2$-distribution (with too large degrees of freedom) is not a good approximation for the distribution of the test statistic even in the limit of large data size.
In summary, we cannot know whether the likelihood ratio test over- or underfits, or chooses the correct order, even if the network is small ($100$ nodes $350$ edges), the ground truth order is small ($\Kgt=2$) and the number of observed transitions is large ($10^7$).

Of the four methods discussed in our work, we find that the Bayesian method performed best.
It lacks the tendency to overfit exhibited by the likelihood ratio test, and it is considerably more data efficient compared to AIC and BIC.
In \cref{fig:order detection comparison} (a), for $\Kgt = 2$ with full knowledge of the network, we find that the likelihood ratio test needs more than $5$ times more data, AIC needs almost $20$ times more data, and BIC needs more than $140$ times more data compared to the Bayesian method where we demand "very strong" evidence.
For $\Kgt = 3$ with full knowledge on the constraint (\cref{fig:order detection comparison} (b)), these differences are even larger: 
The likelihood ratio test needs more than $20$ times more data than Bayesian method demanding "very strong evidence", while AIC needs almost $70$ and BIC needs more than $870$ times more data.
We find that the likelihood ratio test needed marginally fewer data only in the case for the partially known constraint with $\Kgt = 2$ (\cref{fig:order detection comparison} (c)), where it needs $95\%$ of the data needed using Bayes factors. 
For $\Kgt = 3$ (\cref{fig:order detection comparison} (d)) it is again less data-efficient, demanding $30\%$ more data than the Bayesian approach.
We highlight that the proposed Bayesian method is based on an analytical expression for the model evidence, i.e. we do not require Monte Carlo methods for the integration.
Hence, the observed improvements over MLE-based techniques are gained despite the fact that the Bayesian method has the same computational complexity.

We finally comment on potential threats to the validity of our work.
We first note that the overfitting of the likelihood ratio test shown in our experiments was not observed in \cite{scholtes2017network}. 
We suspect that this is due to the mistake in the degrees of freedom calculation, which we mentioned and corrected in \cref{sec:methods}.
This mistake tends to additionally penalize higher orders, thus masking the tendency of the likelihood ratio test to over-fit for moderate data sizes.
Furthermore, we used the argument of \citet{scholtes2017network} to justify the use of MON models as opposed to unconstrained Markov chains.
To ensure that the mistake mentioned above does not challenge the advantages of multi-order models over unconstrained Markov chains, we experimentally reproduced the finding of \cite{scholtes2017network} that multi-order models are more data-efficient. 
The interested reader can find those results in the \cref{sec:Constrained vs unconstrained models}.
Another potential threat to validity could be that our findings depend on the networks that we used as underlying constraints, and which determine the degrees of freedom of the model.
In \cref{sec:Effect of network topology on data efficiency} we checked whether our results qualitatively change under different network constraints and found no evidence for it.
Finally, to illustrate the application of our method to real data on paths (where, however, the ground truth Markov order is unknown), in the \cref{sec:Real-world data} we report the detected Markov order for several data sets that were previously analyzed in \cite{scholtes2017network}.

\section{Conclusion} 
\label{sec:Conclusion}

An increasing volume of categorical sequence data contains large numbers of short, variable-length paths observed in a network, e.g., web users navigating through networks of linked documents, information propagating in social networks, or passengers travelling through transportation networks.
The analysis of such path data requires higher-order network models that capture sequential patterns, while accounting for the constraints imposed by the underlying network.
Learning the optimal order of such higher-order network models is an open problem~\citep{lambiotte2019networks}, which we tackle in this work.
We correct a recently proposed solution based on the likelihood ratio test, and adapt AIC and BIC-based estimators of the Markov order to our specific problem.
We further derive a Bayesian approach to learn an optimal multi-order network model for paths.
Our experimental results show that the proposed Bayesian method considerably outperforms MLE-based methods.
We find that it more reliably learns the correct Markov order of paths, requires less data, and is robust in situations where we have only partial knowledge about the constraints imposed by the underlying network.

Our work opens several perspectives for future research:
First, our formulation allows to include \emph{higher-order constraints} such as, e.g., non-backtracking walks or sequences constrained to complex ``path motifs'', which could improve model selection.
Second, while we have focussed on a flat prior (\cref{eq:prior}), it is an open question how different choices of the prior affect Bayesian model selection.
Third, different from MLE-based approaches, our Bayesian method assigns probabilities to multi-order models with different maximum orders $K$, which can be leveraged for ensemble learning techniques.
Finally, it would be interesting to adapt other Markov order detection techniques mentioned in \cref{sec:background} to multi-order models of paths, e.g. techniques based on surrogate data~\cite{correa2020constrained,pethel2014exact,van1998testing} or conditional mutual information~\cite{papapetrou2013markov}.
We believe that our work serves as a baseline for the development of further methods to learn the Markov order of paths.

The broader impact of our work is due to the growing interest in the role of higher-order interactions in complex networks, i.e. non-dyadic relations that not only involve pairs of nodes \emph{directly} connected via edges.
Data on paths observed in social, technical, and biological networks have become an important source of information on non-dyadic, \emph{indirect} interactions in complex networks.
Neglecting such higher-order interactions limits our ability to analyse rich time-stamped data on networks, and hinders the modelling of complex systems.
Higher-order interactions influence if and how nodes can indirectly influence each other, thus fundamentally challenging our understanding of the \emph{causal topology} of complex systems.
Consequently, the development of higher- and multi-order models for paths in networked systems has become a major focus of the interdisciplinary network science community~\cite{lambiotte2019networks,Battiston2020_Networks}.

Recent works in this field have highlighted opportunities resulting from higher-order generalizations of network analytic methods such as, e.g. random walk models, community detection, centrality measures, node embedding, anomaly detection, or link prediction~\cite{rosvall2014memory,Scholtes2016_HigherOrder,scholtes2017network,Xu2016_Representing,Xu2017_Detecting,Mellor2018_EventGraph,Saebi2019_HONEM}.
However, a major stumbling block is the lack of methods suitable to determine optimal higher-order models for paths that are constrained to a known network topology.
Combining higher-order models of paths studied in network science with model selection and statistical learning, we seek to close this gap.
Apart from enabling network scientists to learn the optimal Markov order of paths in a network, our work contributes to answering the important question for which systems standard network models (of first-order) are sufficient, and in which cases such models are likely to underfit interaction patterns.

\paragraph{Acknowledgements}
Luka Petrovi\'c and Ingo Scholtes acknowledge support by the Swiss National Science Foundation, grant 176938. Ingo Scholtes acknowledges support by the project bergisch.smart.mobility, funded by the German state of North Rhine-Westphalia.
\bibliographystyle{abbrvnat}
\bibliography{model_selection}

\appendix
\section{Marginal likelihood: step by step derivation}
\label{sec:Marginal likelihood: step by step derivation}
In this section we show the step by step derivation of the marginal likelihood, shown in \cref{eq:normalization}.

First we write the definition of the marginal likelihood:
\begin{equation}
    \label{eq:marginal likelihood step 1}
  	p(\data|\prior, K, \network) 
    = 
	\int\limits_{\prod_{h} \probspace{ \Nsuccessors{h}} } 
	    p(\data|\MON{\network}{K}, \prior, K, \network) 
	    p(\MON{\network}{K}|\prior, K, \network) 
	    d\MON{\network}{K} 
\end{equation}
We then substitute the the formulas for likelihood \cref{eq:likelihood} and the prior \cref{eq:prior}:
\begin{equation}
	p(\data|\prior, K, \network) 
	= 
    	\int_{\prod_{h} \probspace{ \Nsuccessors{h}} }  
	        \left[
				\multiplicative
				\prod_{k = 0}^{K}
				\prod_{ h\in \paths{\network}{k}} 
				\prod_{v \in \successors{h}}
				\probHON{h}{v}^\obs{h}{v}
	        \right]
	        \left[
		    	\prod_{k=0}^{K} 
		    	\prod_{h\in \paths{\network}{k}} 
		    	\Dirichlet{\probvecHON{h}}{\hypervec{h}}
	        \right]
	        d\MON{\network}{K}
    \label{eq:marginal likelihood step 2}
\end{equation}
We use the fact that random variables $\probvecHON{h}$ for different $h$ are independent to pull the products outside of the integral.
\begin{equation}
	p(\data|\prior, K, \network) 
	= 
      \multiplicative
      \prod_{k = 0}^{K}
      \prod_{ h\in \paths{\network}{k}} 
      \int_{ \probspace{ \Nsuccessors{h}} }  
        \left[
          \prod_{v \in \successors{h}}
            \probHON{h}{v}^\obs{h}{v}
        \right]
    	\Dirichlet{\probvecHON{h}}{\hypervec{h}}
        d\probvecHON{h}
    \label{eq:marginal likelihood step 3}\\
\end{equation}
We note that the integral above is, by definition, a moment of the Dirichlet distribution.
\begin{equation}
    p(\data|\prior, K, \network) 
	= 
    \multiplicative
      \prod_{k = 0}^{K}
      \prod_{ h\in \paths{\network}{k}} 
        \mathbb{E}_{
          \Dirichlet{\probvecHON{h}}{\hypervec{h}}
        }
        \left[
        {
          \prod_{v \in \successors{h}}
            \probHON{h}{v}^\obs{h}{v}
        }
        \right]
        \label{eq:marginal likelihood step 4}\\
\end{equation}
Therefore we substitute the formula for the moment of the Dirichlet distribution, ($\Betafunc{\vec x}$ denotes the well-known multivariate beta function)
\begin{equation}
    p(\data|\prior, K, \network) 
	= 
    \multiplicative
      \prod_{k = 0}^{K}
      \prod_{ h\in \paths{\network}{k}}
      \frac{
        \Betafunc{\hypervec{h}+\obsvec{h}}
      }{
        \Betafunc{\hypervec{h}}
      }  ,
    \label{eq:marginal likelihood step 5}
\end{equation}
which is the final result.

\section{Exponential penalization of higher orders}
\label{sec:Exponential penalization of higher orders}

The other apriori distribution of the orders of the Markov chain that some researchers use \cite{strelioff2007inferring} is the prior that additionally penalizes higher orders based on the number of the degrees of freedom.
We applied this intuition to the MON models:
\begin{equation}
  \priorK(K) = \frac{\euler^{-\dof{\MON{\network}{K}}}}{\sum_{K'\in \mathbb{K}} \euler^{-\dof{\MON{\network}{K'}}}}.
  \label{eq: exp prior K}
\end{equation}

\begin{figure}[!ht]
  \centering
  \includegraphics[width = \linewidth]{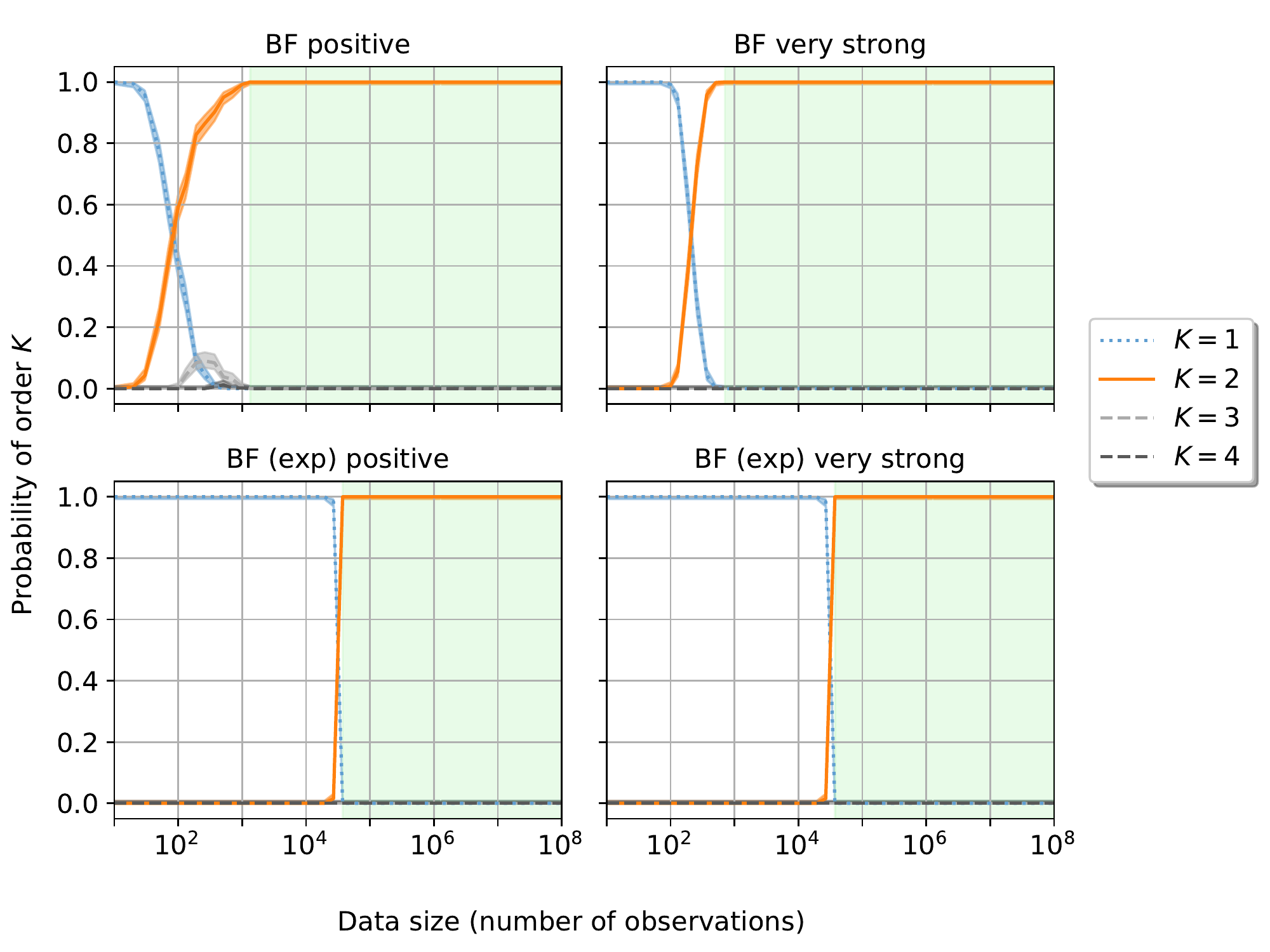}
  \caption{The upper row shows the performance of the Bayes factor mathod with the uniform prior $\priorK$, while the lower row shows the performance of Bayes factor method with the prior $\priorK$ as defined in \cref{eq: exp prior K}.
  }
  \label{fig:exp vs unif prior}
\end{figure}

We use the same experimental setup as for the Fig.~1 in the main paper: $100$ nodes, $350$ edges, paths with ground truth markov order $\Kgt = 2$, and run $500$ independent experiments for each data-size.
In \cref{fig:exp vs unif prior} we show the frequencies of choosing the order $K$ with the Bayes factor method.
We show in the upper row the uniform prior with either ``positive'' or ``very strong'' evidence requirement, and in the lower we show the prior $\priorK$ that penalizes the number of degrees of freedom exponentially, as defined in \cref{eq: exp prior K}.
We obtain that the exponential prior needs more than 50 times more data to detect the correct order.

This behaviour is expected, because the prior is biased towards smaller orders, and we need a lot of evidence to overcome this bias.
In our other experiments, we saw similar behaviour for $\Kgt = 3$, and the partially known constraint.

\section{Efficient determination criterion (EDC)}
\label{sec:Efficient determination criterion (EDC)}

For the sake of transparency, we show here our attempt to adapt the \emph{efficient determination criterion} to the problem of determining the Markov order of paths.
As the results will show, this attempt did not yield a good estimator of the Markov order, which is why we did not include it in the comparison in the main work.

Regarding the EDC for general Markov chains, \citet{zhao2001determination} introduced it and proved its consistency.
Similarly to AIC and BIC, it weighs the log-likelihood of a model with its degrees of freedom.

\begin{equation}
  \label{eq:EDC}
  \EDC{k} \eqdef -2\log{p(s| \MLEmodelMC{k})} +\dof{\MLEmodelMC{k}} c_l
\end{equation}

While \citet{zhao2001determination} gave boundaries for the penalty term $c_l$ for the degrees of freedom, the optimal 
\begin{equation}
  \label{eq:dorea optimal penalty term}
  c_l= \log \log(l)\big/({\Nnodes-1})
\end{equation}
 was derived by \citet{dorea2008optimal}.

\begin{figure}[!ht]
  \centering
  \includegraphics[width = 0.8\linewidth]{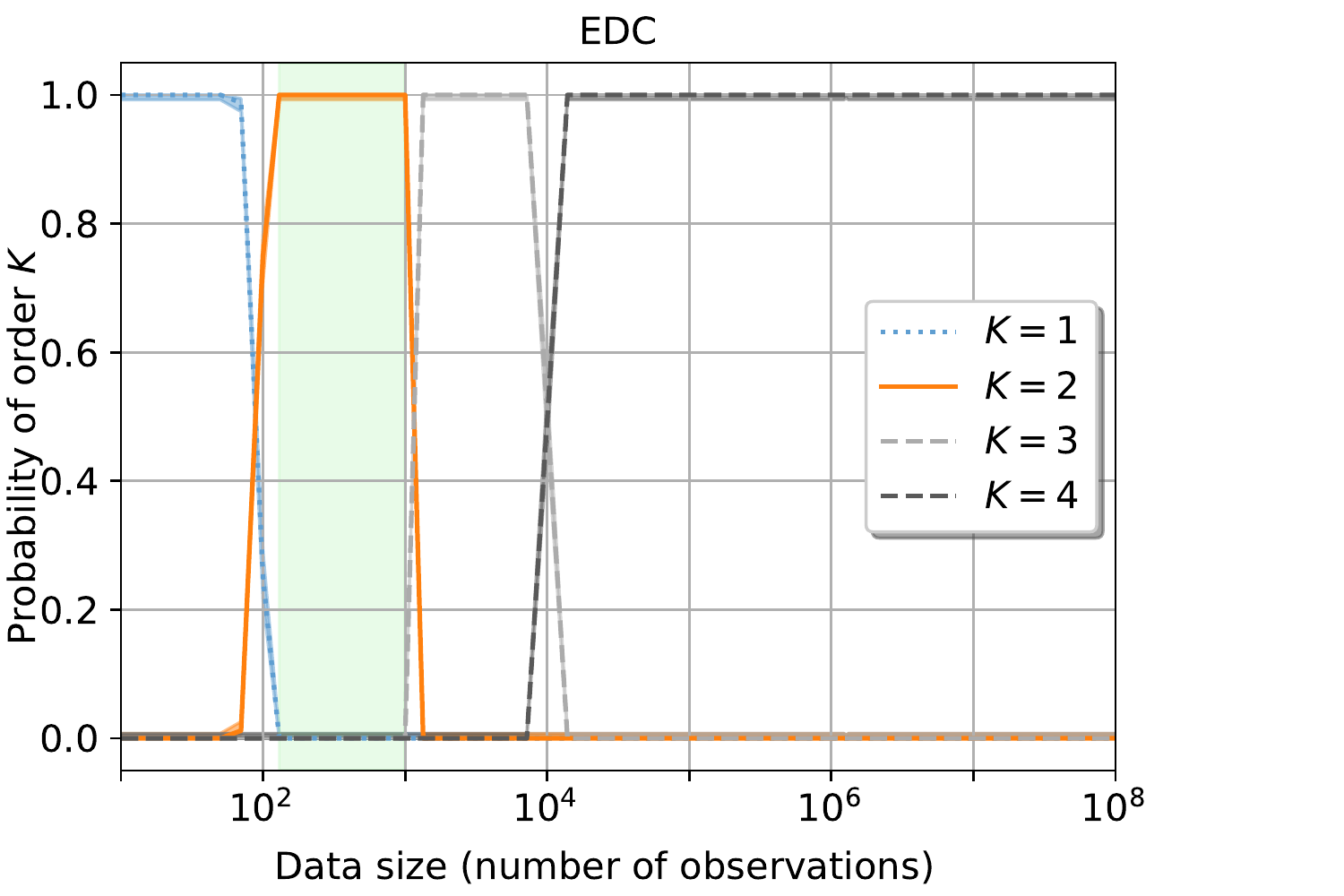}
  \caption{We tested our adaptation of EDC in the same experiment shown in the Fig.~1 in the main manuscript: with $100$ nodes, $350$ edges, $\Kgt = 2$, and $500$ experiments per data-size. We notice that our adaptation of EDC overfitted, and chose the maximal possible order given enough data.
  }
  \label{fig:EDC}
\end{figure}

We adapted this estimator to variable-length paths in a network by simply substituting the Markov chain model for the multi-order network model.
We therefore (re)defined EDC for the context of paths in the following manner:
\begin{equation}
  \label{eq:EDC}
  \EDC{k} \eqdef -2\log{p(s| \MLEmodelMON{\network}{k})} +\dof{\MLEmodelMON{\network}{k}} \frac{\log\log(l)}{{\Nnodes-1}}
\end{equation}
We tested this version of the EDC in the same experiment where we have tested the other methods \cref{fig:order detection comparison}, and obtained the results that we show in \cref{fig:EDC}. 
This adaptation of EDC clearly over-fitted, because with increasing evidence, the detected order also increases.
We leave the further investigation of EDC in the context of paths in a network for future work.

\section{Other results} 
\label{sec:other_results}

\begin{figure}[!h]
  \centering
  \includegraphics[width = 0.9\linewidth]{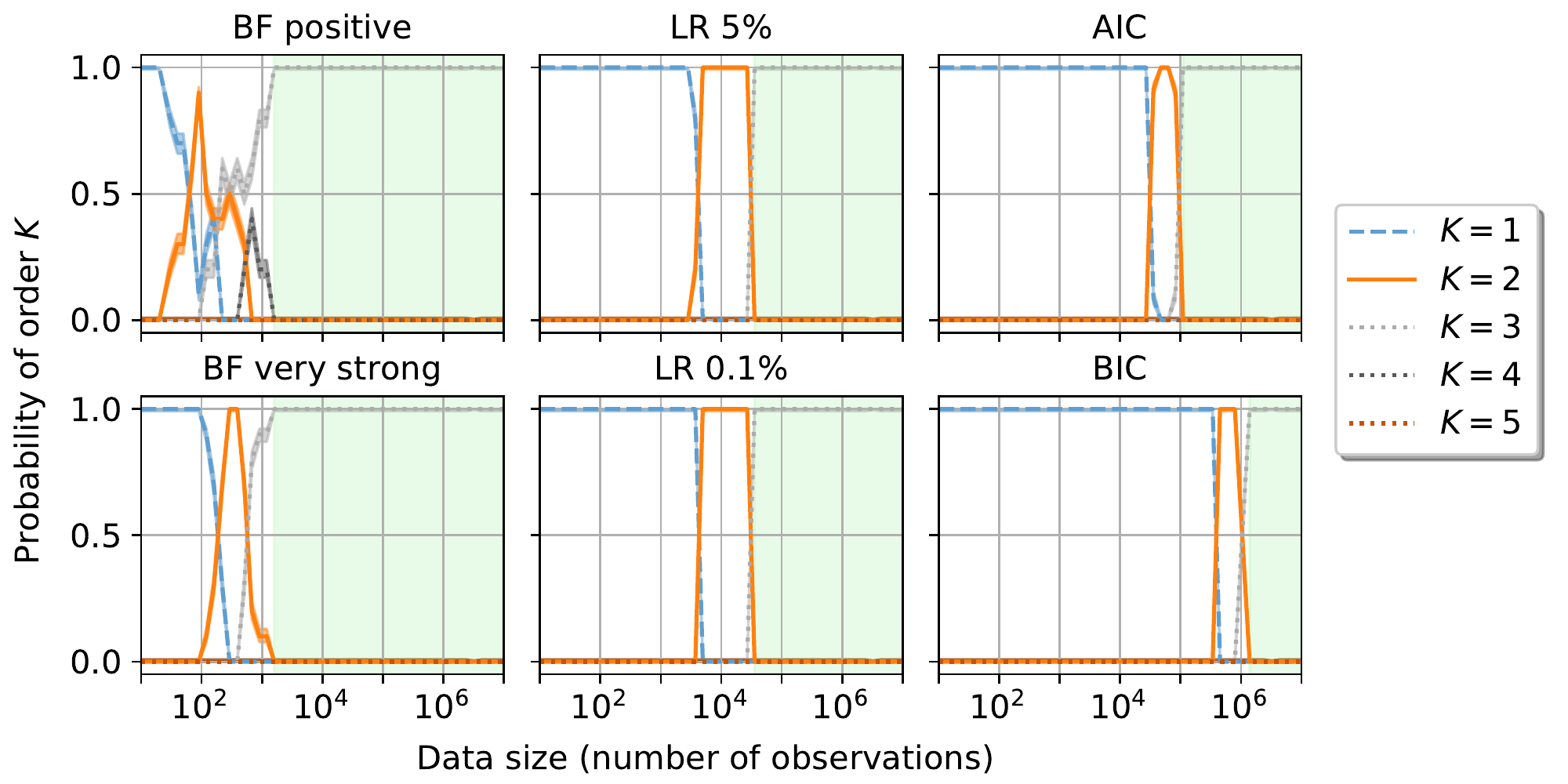}
  \caption{ $\Kgt = 3$
  }
  \label{fig:K3}
\end{figure}

In this section, we show results of experiments identical to the one we presented in the \cref{fig:order detection comparison} (random $G(n,m)$ network as the underlying topology of the process with $100$ nodes and $350$ edges, and $500$ independent experiments per data-size), albeit with the ground truth order $\Kgt = 3$. 
We can see the results in \cref{fig:K3}.
These results are also shown, in the form of data ranges, in Fig.~2 in the main manuscript.
They do not qualitatively differ from the results from $\Kgt = 2$: Bayes factors with the demand of very strong evidence performed best. 

\section{Validity}
\label{sec:validity}
At the end of the discussion of the main manuscript we mentioned two threats to validity of our work that we alleviated by additional experiments.
The first issue is related to the justification of constrained models as opposed to unconstrained models.
The second issue is that we only used a single choice of parameters with $100$ nodes and $350$ edges in our experiment, which raises the concern whether our results generalize to other topologies. 
Addressing these issues, here we present the results of additional experiments.

\subsection{Constrained vs unconstrained models}
\label{sec:Constrained vs unconstrained models}

\begin{figure}[!ht]
  \centering
  \includegraphics[width = \linewidth]{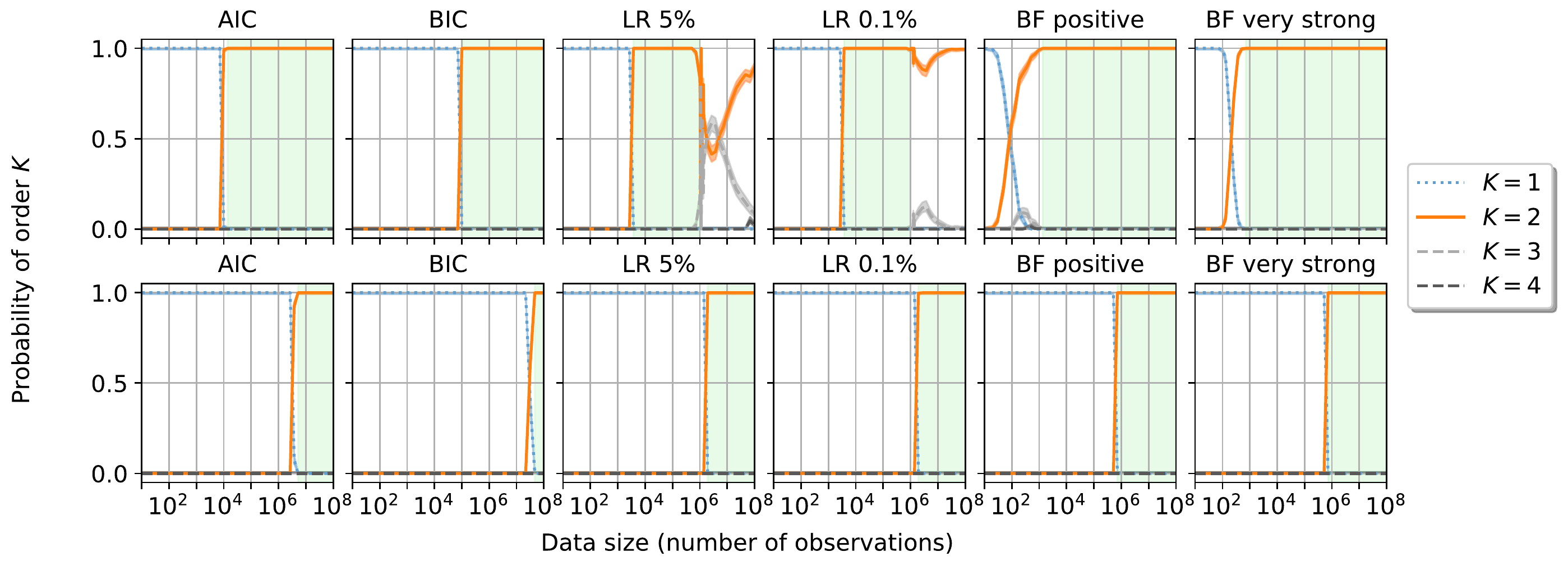}
  \caption{
  Upper row - with constraints, lower row - without constraints (using a fully connected topology as the constraint). 
  $\Kgt = 2$.
  Green area shows the area where the correct order is reliably detected. 
  }
  \label{fig:gt vs no constraint}
\end{figure}

First, we note that, our work relies on the assumption that we do not need to consider models without constraints, which is the main result of \citet{scholtes2017network}.
However, we pointed out a flaw in \cite{scholtes2017network}, in the formula for the degrees of freedom.  
Thus, we checked whether a model selection on unconstrained models showed better data-efficiency than the model selection on constrained models, and present the results in \cref{fig:gt vs no constraint}.
In the top row of \cref{fig:gt vs no constraint} we show the results from \cref{fig:order detection comparison}, and in the bottom, we show the results with the setup that is exactly the same aside that models are unconstrained (as a constraint, we used a fully connected topology, which allows every transition).

We can see in \cref{fig:gt vs no constraint}, as expected from the argument given in the introduction of the main manuscript, that the constrained models are more sensitive (data-efficient) than the unconstrained models that do not assume any apriori knowledge of impossible transitions.

\subsection{Effect of network topology on data efficiency}
\label{sec:Effect of network topology on data efficiency}

In this section we address the question whether our results may be sensitive to the network chosen for the ground truth path-generation.

\begin{figure}[!ht]
  \centering
  \includegraphics[width = 0.6\linewidth]{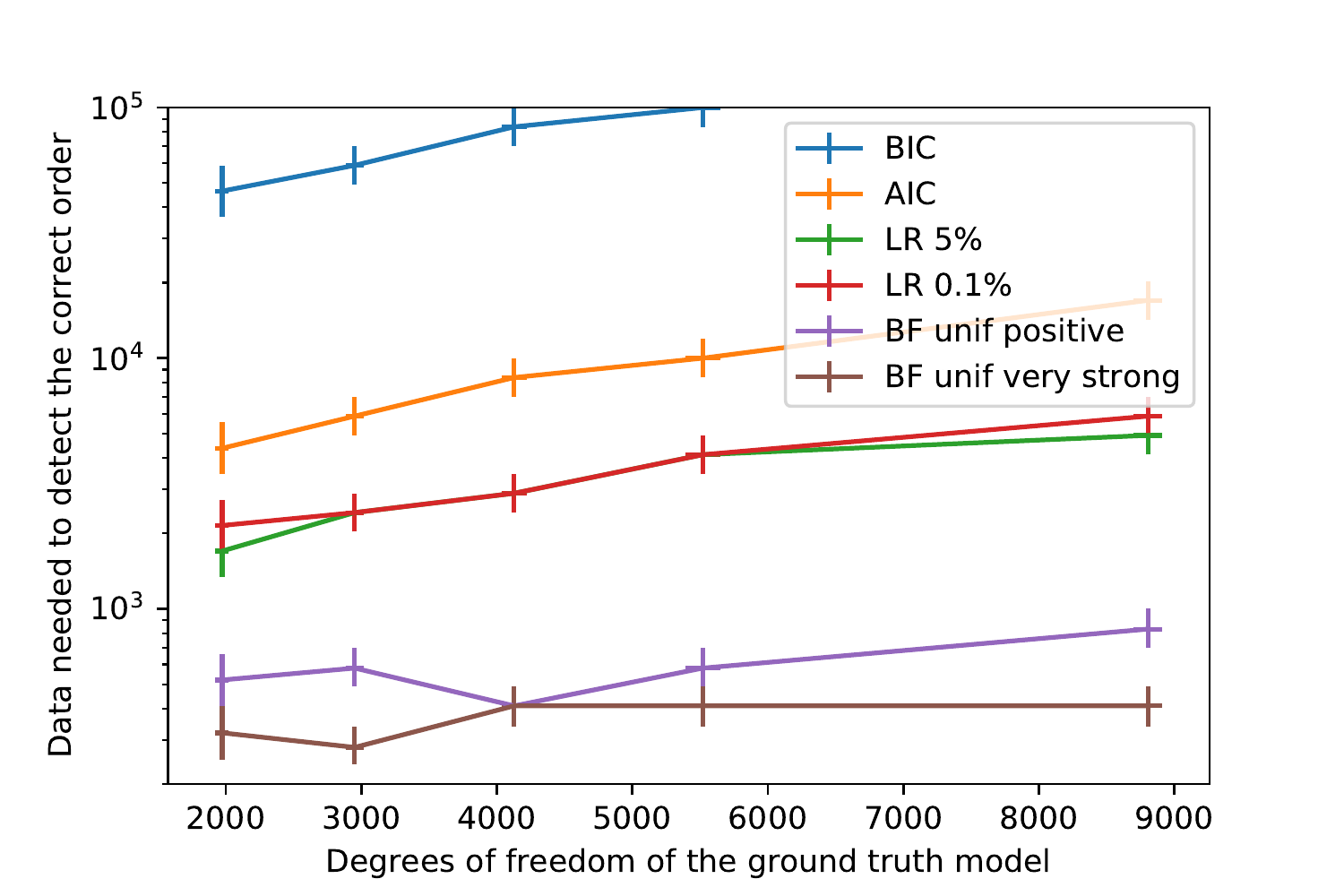}
  \caption{ Least data needed to detect the correct order using different methods, for varying number of the degrees of freedom.
  }
  \label{fig:network sizes}
\end{figure}

We vary the number of edges $m$ in the Erd\"os-R\'enyi $G(n,m)$ model that generates the underlying topology of the ground truth model that generates paths.
We ran experiments like we have shown in the Fig.~1. in the main manuscript with different numbers of edges: $200, 250, 300, 350, 450$.
For each case, we determine the lower boundary of the data range where the method guessed the correct Markov order of paths every time i.e. the lower boundary of the light-green area in Fig.~1 of the main manuscript, and plot the dependency of the lower boundary to the degrees of freedom of the model.
We found the number of degrees of freedom to be the most meaningful to show, because it encompasses the number of nodes and edges of the underlying topology, and the ground truth Markov order of the paths, all at the same time.
The dependency is shown in \cref{fig:network sizes}.
The results show no evidence that the evidence should be rejected.

\section{Real-world data}
\label{sec:Real-world data}

To demonstrate the application of the methods in practical scenarios, we show the Markov order detected using all methods discussed in the paper in three different empirical data sets.

The first data set represents the click stream paths in the Wikispeedia game (WIKI) \cite{west2012human}.
Users are instructed to navigate from a given starting article to a given goal using only the hyperlinks connecting the pages.
For the network constraint $\network$, we used the hyperlink network between the Wikipedia articles that the users navigated on.
This graph has 4592 nodes and 239~764 directed edges.
There are in total 76~192 observed paths, with 475~889  observations of transitions.

Second data-set consists of flight routes in the network of US airports measured in 2001 (FLY) \cite{FLdata}.
For the network constraint $\network$, we assumed that every possible connection between the airports had at least one passenger using it.
This graph has 175 nodes and  1598 edges.
There are in total 286~810 
observed paths, with 1.2 million observations of transitions.

Third data set captures passenger trajectories in the network of London subway stations (TUBE) \cite{LTdata}.
For the network constraint $\network$, we took the map of the London tube.
The paths were inferred from the origin-destination data, as the shortest path between the origin and the destination. 
This graph has 276 nodes and 666 directed edges.
There are around 4.3 million observed paths, with 34 million observations of transitions.

\begin{table}[!h]
  \caption{Detected orders in real data-sets}
  \label{sample-table}
  \centering
  \begin{tabular}{cccccc}
    \hline \\
    Data     & Max order tested & AIC 	& BIC  	& LR test 	& BF "very strong"		\\
    \hline \\
    WIKI 	 &	4				& 1 	& 1		& 1			& 1  					\\
    FLY      &	6				& 2 	& 1		& 2			& 4					  	\\
    TUBE     &	14				& 5 	& 4		& 6			& 12  					\\
    \hline\\
  \end{tabular}
\end{table}

We notice that the detected order for WIKI data differs from the one detected in \cite{scholtes2017network}, and attribute this difference to the fact that they used paths between only the 100 most visited nodes, ignoring the rest of the articles.
Thus they have used only the part of the network for which there is enough evidence to assert higher-order correlations.
The reason why we did not follow their procedure is two-fold: first, we wanted to see whether there would be enough evidence to detect the higher-order correlations even when we consider the whole network, and second, we wanted to test whether we could use the methods in a larger real-world network with a reasonable processing time.
While there wasn't enough evidence to assert a higher order, we find that the methods are sufficiently scalable that we needed a few hours on a consumer-grade machine to test up to fourth order.

The orders, detected with likelihood ratio test, in the other two data-sets coincided with \cite{scholtes2017network} despite the mistake in the calculation of the degrees of freedom in \cite{scholtes2017network}.
We notice that the Bayes factor is indeed more sensitive, detecting larger optimal orders than the other methods.
These results confirm that our method is relevant to infer the optimal Markov order in real-world path data.  

\end{document}